# COMBINATION OF UPPER AND LOWER PROBABILITIES

**Jose E. Cano,**
*Departamento de Ciencias de la Computación e I.A.
Universidad de Granada.
18071 Granada, Spain.*

**Serafín Moral,**

**Juan F. Verdegay-López**

## Abstract

In this paper, we consider several types of information and methods of combination associated with incomplete probabilistic systems. We discriminate between 'a priori' and evidential information. The former one is a description of the whole population, the latest is a restriction based on observations for a particular case. Then, we proposse different combination methods for each one of them. We also consider conditioning as the heterogeneous combination of 'a priori' and evidential information. The evidential information is represented as a convex set of likelihood functions. These will have an associated possibility distribution with behavior according to classical Possibility Theory.

## 1 INTRODUCTION

In Probability Theory the main method of incorporating new information is through conditioning. In general, there is no doubt about how to represent initial information and how to update it in the light of new observations. However, when we work with probabilistic intervals, there is a bit of a mess. Several formulas of combining and conditioning are available, the main problem being which formula or method to use in a particular case. The reason for this is the lack of a firm semantic basis.

In this paper we present a method of tackling incomplete probabilistic information, which tries to avoid this kind of ambiguity. The main feature is the clear distinction between evidential and 'a priori' information. Evidential information will be represented as a convex set of likelihood functions with an associated possibility distribution. The interpretation of these possibilities will have a probabilistic basis. However their behavior will be very similar to classical possibility distributions (Dubois, Prade 1988; Zadeh 1978).

We shall discriminate between the combination of evidential information and the combination of 'a priori' information. Conditioning will be a kind of heterogeneous combination: 'a priori' and evidential information. The result is called 'a posteriori' information. This 'a posteriori' information is different from that resulting from applying known formulas of calculating conditional information: Dempster conditioning (Dempster 1967) and upper and lower conditioning (Dempster 1967; Campos, Lamata, Moral 1990; Fagin, Halpern 1990).

The problem of applying Dempster-Shafer Theory of Evidence (Dempster 1967; Shafer 1976) to upper and lower probabilities is that there is only one way of combining information, the so called Dempster rule of combination. Furthermore, the most used conditioning is Dempster conditioning (Dempster 1967; Moral, Campos 1990) which is a particular case of Dempster rule. So all information is combined in an homogeneous way. This gives rise to cases in which Dempster rule seems reasonable and cases in which the results are not very intuitive. In general, Dempster conditioning produces very narrow intervals, if upper and lower probabilities interpretation is considered (Pearl 1989).

## 2 'A PRIORI' INFORMATION

Let $X$ be a variable taking values on a finite set $U = \{u_1, ..., u_m\}$. An 'a priori' information about $X$ is a convex set of probabilities,

$$\mathcal{H} = \left\{ \sum_{i=1}^{n} \alpha_i P_i \mid \sum_{i=1}^{n} \alpha_i = 1 \right\}$$

where $\{P_1, P_2, ... P_n\}$ is a finite set of probabilities on $U$. In general, this information is applicable under a determined set of conditions $Co$ under which $X$ takes its values. The meaning is that one element of $\mathcal{H}$ is the true probability distribution associated with $X$, under conditions $Co$. Given a set of possible probabilities $\mathcal{C} = \{P_1, P_2, ... P_n\}$, we may associate a convex set of probabilities with it, its convex hull

$$\overline{\mathcal{C}} = \left\{ \sum_{i=1}^{n} \alpha_i P_i \mid \sum_{i=1}^{n} \alpha_i = 1 \right\}$$

$\mathcal{C}$ and $\overline{\mathcal{C}}$ may be considered for the same experiment with different interpretations. For example, let us consider that we have two urns, $U_1$ and $U_2$. $U_1$ has 99 red balls and one black. $U_2$ has one red ball and 99 black ones. If we pick up a ball randomly from one of the two urns then, we have two possible probabilities, $\mathcal{C} = \{P_1, P_2\}$, for the color of the ball, one for each urn. However, if the experiment is to select an urn and then pick a ball, the frequencies of the colors will be given by one probability $\alpha P_1 + (1 - \alpha)P_2$, where $\alpha$ is the probability of selecting $U_1$, and $(1 - \alpha)$ the probability of selecting $U_2$. As $\alpha$ is unknown, we have a convex set of probabilities $\overline{\mathcal{C}}$. We will always consider the second interpretation. That is, that we have a previous experiment consisting in randomly selecting one of the possible probability distributions. Therefore a set of probability distributions will be equivalent to its convex hull.

Above assumption may have some problems. For example, assume that we have two variables, $X_1$ and $X_2$, which are known to be independent. For the first variable we have the convex set of possible probabilities $\mathcal{C}_1 = \overline{\{P_1, Q_1\}}$, and for the second variable, the convex set $\mathcal{C}_2 = \overline{\{P_2, Q_2\}}$. Given these conditions, the possible set of probabilities for the variable $X = (X_1, X_2)$ is

$$\mathcal{C} = \{p \mid p(a_1, a_2) = h_1(a_1)h_2(a_2), h_1 \in \mathcal{C}_1, h_2 \in \mathcal{C}_2\}$$

However, this set $\mathcal{C}$ is not neccesarily convex. We may have:

- $P_1.P_2, Q_1.Q_2 \in \mathcal{C}$

- Sometimes the convex combination

$$\alpha(P_1.P_2) + (1 - \alpha)(Q_1.Q_2)$$

is not equal to a product such as

$$(\beta P_1 + (1 - \beta)Q_1).(\gamma P_2 + (1 - \gamma)Q_2),$$

that is, an element from $\mathcal{C}$.

This has been considered also by B. Tessem (1989), in a slightly different problem: The induced set of probabilities in $X_2$, from a convex set of 'a priori' distributions on $X$, and a convex set of conditionals. It is shown that this set is not necessarily convex.

We shall always represent uncertainty as a convex set of probabilities. Then, in situations like this, we shall do an approximation of the possible set of probabilities calculating its convex hull. From a practical point of view, this approximation is equivalent to assuming that the selection of probabilities for $X_1$ and $X_2$ may be done in a dependent way. That is, with probability $\alpha$ we may choose $P_1$ for $X_1$ and $P_2$ for $X_2$; and with probability $1 - \alpha$ it is possible to choose $Q_1$ for $X_1$ and $Q_2$ for $X_2$. This does not imply that the variables $X_1$ and $X_2$ are dependent. In fact, the individual probabilities are combined by multiplication ($P_1.P_2$ and $Q_1.Q_2$). Only the probability distributions are selected on a dependent way. In a particular case, it may occurs that the selection of probabilities is independent. But then, we are adding some extra probabilities. From this point of view, we are losing some information, but we lose it for the sake of simplicity. Convex sets are more manageable than general sets.

The combination of 'a priori' convex sets of probability distributions has been considered in Campos, Lamata, Moral (1988). If we have two 'a priori' convex sets of probabilities $\mathcal{C}_1$ and $\mathcal{C}_2$ given for the same set of conditions, then the conjunction will be the intersection of the convex hulls, $\overline{\mathcal{C}}_1 \cap \overline{\mathcal{C}}_2$. The disjunction will be the convex hull of the union: $\overline{\mathcal{C}_1 \cup \mathcal{C}_2}$.

This kind of combination is the one applied in situations like the following: We know that if we pick up a ball from this urn the probability of being red is between 0.75 and 0.85, and from other source we obtain that this probability is between 0.8 and 0.9. Then we may apply this combination (conjunction in this case) and deduce that the probability is between 0.8 and 0.85.

If $\mathcal{C}_1$ and $\mathcal{C}_2$ are convex sets of probabilities, but relative to different contexts $Co_1$ and $Co_2$, then no 'a priori' information may be deduced in the context $Co_1 \cup Co_2$ (both conditions are verified), except if one of the probabilities is degenerated. The following situation is perfectly possible:

- $C_1 = \{p_1\}$ in conditions $Co_1$, where

$$p_1(a_1) = 0.99, \ p_1(a_2) = 0.01$$

- $C_2 = \{p_2\}$ in conditions $Co_2$, where

$$p_2(a_1) = 0.99, \ p_2(a_2) = 0.01$$

- $C_3 = \{p_3\}$ in conditions $Co_1 \cup Co_2$, where

$$p_3(a_1) = 0, \ p_3(a_2) = 1$$

However, if we have the following information,

- I1: "Most of Computer Science students (CS) are single (S)"
- I2: "Most of young people (Y) are single (S)"

Where I1 and I2 may be translated in the probabilistic information,

- Under conditions $Co_1 = \{CS\}$

$$p_1(S) = 0.99 \quad p_1(\neg S) = 0.01$$

- Under conditions $Co_2 = \{Y\}$

$$p_2(S) = 0.99 \quad p_2(\neg S) = 0.01$$

and nothing is known about the probability of being single under conditions {CS,Y}, then common sense says that in this case it would not be wrong to assume that "Most young people studing Computer Science are single". It migth occur that young Computer Science students are a rare combination and most of them are married. But if nothing is said about this, then it may be considered that there are not strange interactions and that under conditions $Co_1$ and $Co_2$ we may use $C_3 = C_1 \cap C_2 = \{p_1\} = \{p_2\}$.

In short, to do the combination of 'a priori' probabilitic convex sets is neccesary to determine whether they are given in the same context. In shuch a case, we calculate the conjunction by the intersection of convex sets and the disjunction by the convex hull of the union. If the sets of probabilities are given under two different contexts, then nothing can be said about the combination. However, when there is no more available information, then the former combination could be considered by default, but taking into account that this is an additional assumption we make about the problem.

Very often, 'a priori' information is given by means of probability intervals or probability envelopes. A probability envelope is a pair of applications

$$l, u : \mathcal{P}(U) \longrightarrow [0,1]$$

such that there exist a family $\mathcal{P}$ of probability measures verifying

$$l(A) = \inf_{P \in \mathcal{P}} P(A) \quad u(A) = \sup_{P \in \mathcal{P}} P(A)$$

It is clear that given a set of probabilities, $C_1$, we may associate with it a probability envelope. However a probability envelope $(l, u)$ may be defined from different sets of probabilities. But, in every case, there is always a maximal family given by

$$\mathcal{P} = \{P | l(A) \leq P(A) \leq u(A), \forall A \subseteq \mathcal{U}\}$$

If $C$ is a set of probabilities and we calculate the associated envelope $(l, u)$, this envelope is equivalent to a maximal family $\mathcal{P}$. Always, we have $C \subseteq \mathcal{P}$. Then, if we transform a set $C$ on an envelope we may consider that some information is lost (there are more probabilities being possible).

## 3 EVIDENTIAL INFORMATION

### 3.1 LIKELIHOOD FUNCTIONS

Assume that we do not have an 'a priori' information about X, but we have observed O, and we have a family of conditional probabilities,

$$P(O|X = a_i), \ a_i \in U.$$

Taking into account that nothing is known about 'a priori' probabilities of $a_i$ are not known then nothing can be said about 'a posteriori' probabilities of $a_i$ after observing $O$, with the exception that if $P(O|a_i) = 0$ then we can conclude that $a_i$ is impossible. Cosider the following example: We have $U = \{a_1, a_2\}$ and

$$P(O|a_1) = 1, P(O|a_2) = 0.001$$

Then after observing $O$, we may have $p(a_2|O) = 1$ $(p(a_2|O) = 0)$ if the unknown 'a priori' probability was $p(a_2) = 1$ $(p(a_2) = 0)$. However, it is clear that after observing O, $a_2$ should be considered less possible than $a_1$. In conclusion the information provided by O can not be represented by probabilities or interval probabilities. We shall do as in Classical Non-Bayesian Statistics and say that $O$ defines a likelihood function, $l_o$, on $U$, which is a mapping from $U$ on the interval $[0, 1]$, given by

$$l_o(a_i) = P(O|X = a_i), \ a_i \in U.$$

This likelihood may be interpreted as a possibility distribution, $\pi_o$, which is not neccesarily normalized (Smets 1982).

The possibility measure associated with $\pi_o$ is defined (Zadeh 1978) as a mapping

$$\Pi_o : \mathcal{P}(U) \longrightarrow [0,1]$$

given by

$$\Pi_o(A) = \max_{a \in A} \pi_o(a)$$

The following proposition relates a possibility measure with probability bounds.

*Proposition.-* Given a possibility measure $\Pi_o$ on $U$, then

$$\Pi_o(A) \geq P(O \cap A)$$

and these bounds are optimal under information $O$.

*Proof.-*

$$P(O \cap A) = \sum_{a_i \in A} P(O \cap \{a_i\}) =$$

$$\sum_{a_i \in A} P(O|a_i).p(a_i) = \sum_{a_i \in A} \pi_o(a_i).p(a_i)$$

Now, taking into account that,

$$\sum_{a_i \in A} p(a_i) = P(A) \leq 1,$$

we get the required inequality,

$$P(O \cap A) \leq \max_{a_i \in A} \pi_o(a_i) = \Pi_o(A).$$

The bounds are optimal in the sense that being

$$\Pi_o(A) = \max_{a_i \in A} \pi_o(a_i)$$

then, if

$$\max_{a_i \in A} \pi_o(a_i) = \pi_o(a_k), \quad a_k \in A$$

we may consider the 'a priori' probability

- $p(a_k) = 1$,

- $p(a_i) = 0$, *otherwise*

With this 'a priori' probability,

$$P(O \cap A) = p(a_k)P(O|a_k) = \Pi_o(A),$$

that is, equality is given. ∎

These bounds are not associated with conditional probabilities, $P(.|O)$, but with probabilities of consistency with information $O$. To consider real conditional probabilities we have to divide by $P(O)$, but this probability is unknown, and we only have an upper bound $\Pi_o(U)$. The normalization by this value may be considered as an upper relative degree of consistency,

$$g_o^*(A) = \frac{\Pi_o(A)}{\Pi_o(U)}$$

From this upper value, we may define the lower limit as

$$g_{o*}(A) = 1 - g_o^*(\overline{A}).$$

## 3.2 CONVEX SETS OF LIKELIHOOD FUNCTIONS

On the other hand, it is possible that the exact values of conditional probabilities are not known. For example, we only have probability intervals

$$b_i \leq P(O|X = a_i) \leq c_i$$

In this case, observation $O$ does not define only one likelihood function, but a convex set of likelihoods, those verifying $a_i \leq l(a_i) \leq c_i$. This convex set will be called the evidential information associated with $O$ and denote it by $E_o$.

An evidential information also has an associated possibility distribution,

$$\pi_o(a_i) = \max_{l \in E_o} l(a_i)$$

This possibility distribution also verifies a similar proposition to the above, relative to probability bounds. In the same way, we may associate the pair of lower-upper measures $(g_{o*}, g_o^*)$.

With the same reasoning as in 'a priori' information, it will be considered that a set of likelihood functions, $E$, is equivalent to its convex hull, $\overline{E}$. A special likelihood function is the null likelihood, $l_N$, defined as

$$l_N(a_i) = 0, \quad \forall a_i \in U$$

This likelihood function comes from an observation, $O$, for which

$$P(O|a_i) = 0$$

are possible conditional probabilities (there may be another possible conditional probabilities defining other likelihood functions associated with observation $O$).

It is clear that after observing $O$, these conditional probabilities are impossible, because we have

$$P(O) = \sum_i P(O|a_i)p(a_i) = 0$$

Then $l_N$ has to be considered as an impossible likelihood. It could be thought that when $l_N$ appears it would be better to remove it. In fact, the removing of something impossible must not change our state of mind. But for the same reason the inclusion of $l_N$ should not have any effect in our final results. This will happen in our model: never the final probability intervals will chage because of the inclusion or elimination of $l_N$.

Taking the above reasons into account we shall extend our original equivalence relation among sets of likelihood functions, considering that two sets, $E_1$ and $E_2$, are equivalent if and only if $\overline{E_1 \cup \{l_N\}} = \overline{E_2 \cup \{l_N\}}$. That is, if previously including the null linkelihood their convex hulls are equal. The effect of this equivalence relation will be that if we have a likelihood, $l$, we do not have to consider any likelihood $\alpha.l$ (where $\alpha \leq 1$). This is not strange. Having $l$, the likelihood $\alpha.l$ ($\alpha \leq 1$) defines the same relative possibilities, but with a lower normalization factor.

In the following, the convex set $\overline{E \cup \{l_N\}}$ will be denoted as $C(E)$.

The disjunction and conjunction of evidential information are defined in an analogous way to the disjunction and conjunction of 'a priori' information.

If $E_1$ and $E_2$ are two sets of evidential information, the disjunction, $E_1 \vee E_2$, is defined as $C(E_1 \cup E_2)$. The conjunction, $E_1 \wedge E_2$, is defined as the intersection of the convex hulls: $C(E_1) \cap C(E_2)$.

Here is important to distinguish between the conjunction of evidential information, $C(E_1) \cap C(E_2)$ and the evidential observation associated with the conjunction of two observations $O_1$ and $O_2$, $E_{O_1 \wedge O_2}$. The first is applied when we know that $E_1$ and $E_2$ are two convex sets of likelihood functions associated with the same observation and is performed calculating $C(E_1) \cap C(E_2)$. The last, when we have two sets of likelihoods corresponding to two observations, $O_1$ and $O_2$, and we want to calculate the evidential information associated with $O_1 \wedge O_2$. The same is applied for the disjunction.

For the calculus of evidential information $E_{O_1 \wedge O_2}$ from the evidential information $E_{O_1}$ and $E_{O_2}$, we have to calculate the possible values for the probability $P(O_1 \wedge O_2|a_i)$ from the values of $P(O_1|a_i)$ and $P(O_2|a_i)$. The only thing we can say is that

$$\max\{0, P(O_1|a_i) + P(O_2|a_i) - 1\} \leq P(O_1 \wedge O_2|a_i) \leq$$
$$\leq \min\{P(O_1|a_i), P(O_2|a_i)\}$$

then we have to consider in $E_{O_1 \wedge O_2}$ all the likelihood functions, $l$, verifying

$$\max\{0, l_1(a_i) + l_2(a_i) - 1\} \leq l(a_i) \leq$$

$$\leq Min\{l_1(a_i), l_2(a_i)\}$$

where $l_1 \in E_{O_1}, l_2 \in E_{O_2}$.

For the associated possibility distribution, we get

$$\pi_{O_1 \wedge O_2}(a_i) = \min\{\pi_{O_1}(a_i), \pi_{O_2}(a_i)\}$$

that is, the same formula as in classical Possibility Theory (Dubois, Prade 1988), but without assuming any additional assumption of coherence or compatibility of observations.

If we assume that $O_1$ and $O_2$ are conditionally independent given the value of $X$ then we get:

$$\pi_{O_1 \wedge O_2}(a_i) = \pi_{O_1}(a_i).\pi_{O_2}(a_i).$$

Another assumptions may be make to obtain different combination formulas for operations on observations.

## 4 COMBINATION OF 'A PRIORI' AND EVIDENTIAL INFORMATION

Here, it is considered the combination of a convex set of 'a priori' probabilities, $C$, with an evidential information, $E$. The method is a generalization of Bayes Theorem and is based on the formula of conditioning in Moral, Campos (1990). The generalization given here is different from the one given by Smets (1978, 1981). The main difference comes from the fact that we assume that an observation defines a consonant evidence (a possibility) and Smets works with general evidential information. Also, in our approach, we distinguish between evidential and 'a priori' information using different methods according to the particular situation.

The combination of a probability measure $p$ about the values of $X$ and a possibility associated with observation $O : \pi_o(a_i) = p(O|X = a_i)$, is given by the function,

$$h(a_i) = p(a_i).\pi_o(a_i).$$

We shall denote this function $h$ by $p \times \pi_o$. After normalization, h determines the values of conditional probability,

$$p(X = a_i|O) = \frac{h(a_i)}{\sum_j h(a_j)}$$

The normalization factor may be considered as the likelihood of the 'a priori' probability given O. A very small likelihood of h may make us suspect the initial values of probability and therefore, the resulting conditional probabilities. Furthermore, in this case these values will be very sensitive to the lack of accuracy of initial probabilities.

The above expression is precisely Bayes formula, developed in two stages: first combination and after normalization. In a similar way, we shall define the combination of an 'a priori' convex set of probability distributions and an evidential information $E_o$ as the set,

$$H = C(\{p \times \pi | p \in \mathcal{C}, \pi \in E_o\}) = \mathcal{C} \otimes E_o$$

As before, and by analogous reasons we shall assume that two combination sets, $H_1$ and $H_2$, are equivalent if $C(H_1) = C(H_2)$.

To assign probability intervals (Moral, Campos 1990) to the set $H$ we select the extreme points of $H$, $h_1, ..., h_n$. Each $h_k$ different from the null function can define a probability value, normalizing it by its likelihood,

$$p_k(X = a_i | O) = \frac{h_k(a_i)}{\sum_j h_k(a_j)}$$

We could now calculate the upper and lower probability values by means of the expression,

$$t_o^*(A|O) = \max_k P_k(A|O),$$

$$t_{o*}(A|O) = \min_k P_k(A|O)$$

This is equivalent to upper-lower conditioning (Dempster 1967; Fagin, Halpern 1990). But in this method there is some missing information. In effect, given that $h_k$ is the true combination, then in this case, the probability of observation $O$ is $\sum_j h_k(a_j)$. According to our definition of possibility, this defines a possibility about the combination functions and the corresponding conditional probabilities, given by

$$\pi(p_k(.|O)) = \sum_j h_k(x_j)$$

That is, we do not only have a set of conditional probabilities, we also have a possibility about them. This possibility also defines upper and lower probabilities: If $D \subseteq \{p_1(.|O), ..., p_m(.|O)\}$, then

$$g_o^*(D) = \frac{\Pi(D)}{\max_k \pi(p_k(.|O))},$$

$$g_{o*}(D) = 1 - g_o^*(\overline{D}).$$

Now, we define the upper and lower conditional intervals in the following way. The upper and lower values of $B$ given observation $O$, are calculated by means of Choquet integral (Choquet 1953) of conditional probabilities with respect to the measures $g_o^*$ and $g_{o*}$, respectively:

$$P_o^*(B) = I(P_k(B|O)/g_o^*),$$

$$P_{o*}(B) = I(P_k(B|O)/g_{o*}),$$

these integrals being defined in the following way,

$$I(h/g) = \int_0^\infty g(H_\alpha) d\alpha$$

where

- $h$ is a function $h : U \longrightarrow R_o^+$
- $H_\alpha = \{x \in U | h(x) \geq \alpha\}$
- $g$ is a monotone fuzzy measure (non necessarily additive (Sugeno 1974)).

It is important to point out that the result of the combination is the set $H$, not the intervals. For example, we may have the same intervals coming from pure possibilistic information or a convex set of probability distributions. However, after combining each one of them with new information, the intervals may become very different. The following example illustrates these ideas.

*Example.*- Let us consider a set $U = \{1, 2, 3\}$ and an obsevation $O_1$ such that

$$p(O_1|1) = 1, \quad p(O_1|2) = 0.5, \quad p(O_1|3) = 0.2$$

The intervals defined by this observation are

$$\begin{array}{ll} \emptyset \longrightarrow [0,0] & \{1\} \longrightarrow [0.5, 1] \\ \{2\} \longrightarrow [0, 0.5] & \{3\} \longrightarrow [0, 0.2] \\ \{1,2\} \longrightarrow [0.8, 1] & \{1,3\} \longrightarrow [0.5, 1] \\ \{2,3\} \longrightarrow [0, 0.5] & \{1,2,3\} \longrightarrow [1, 1] \end{array}$$

Now assume that we have a convex set of probability distributions, $\mathcal{C}$, with extreme points

|       | 1   | 2   | 3   |
|-------|-----|-----|-----|
| $p_1$ | 1   | 0   | 0   |
| $p_2$ | 0.5 | 0.5 | 0   |
| $p_3$ | 0.5 | 0.3 | 0.2 |
| $p_4$ | 0.8 | 0   | 0.2 |

The intervals associated with it are the same as before. However the information is different and it is combined in a different way. Assume now that we have observation $O_2$ and

$$p(O_2|1) = 0.1, \quad p(O_2|2) = 0.6, \quad p(O_2|3) = 1$$

If we assume that $O_1$ and $O_2$ are conditionally independent given the value of $X$, then the conjunction of these two observations gives rise to the following possibility and intervals.

$$\pi_{o_1 \wedge o_2}(1) = 0.1 \quad \pi_{o_1 \wedge o_2}(2) = 0.3 \quad \pi_{o_1 \wedge o_2}(2) = 0.2$$

$$\emptyset \longrightarrow [0,0] \qquad \{1\} \longrightarrow [0, 0.33]$$
$$\{2\} \longrightarrow [0.33, 1] \qquad \{3\} \longrightarrow [0, 0.67]$$
$$\{1,2\} \longrightarrow [0.33, 1] \qquad \{1,3\} \longrightarrow [0, 0.67]$$
$$\{2,3\} \longrightarrow [0.67, 1] \qquad \{1,2,3\} \longrightarrow [1, 1]$$

If we combine observation $O_2$ with the convex set $\mathcal{C}$, we get the convex set $H$ with extreme points

|       | 1    | 2    | 3   |
|-------|------|------|-----|
| $h_1$ | 0    | 0    | 0   |
| $h_2$ | 0.1  | 0    | 0   |
| $h_3$ | 0.05 | 0.3  | 0   |
| $h_4$ | 0.05 | 0.18 | 0.2 |
| $h_5$ | 0.08 | 0    | 0.2 |

The corresponding normalized probabilities and possibilities are

| 1         | 2         | 3         | $\pi$ |
|-----------|-----------|-----------|-------|
| undefined | undefined | undefined | 0     |
| 1         | 0         | 0         | 0.1   |
| 0.14      | 0.86      | 0         | 0.35  |
| 0.12      | 0.42      | 0.46      | 0.43  |
| 0.29      | 0         | 0.71      | 0.28  |

The intervals, calculated using Choquet integral are,

$$\emptyset \longrightarrow [0,0] \qquad \{1\} \longrightarrow [0.12, 0.40]$$
$$\{2\} \longrightarrow [0.15, 0.78] \qquad \{3\} \longrightarrow [0.09, 0.63]$$
$$\{1,2\} \longrightarrow [0.37, 0.91] \qquad \{1,3\} \longrightarrow [0.22, 0.85]$$
$$\{2,3\} \longrightarrow [0.60, 0.88] \qquad \{1,2,3\} \longrightarrow [1,1]$$

which are really different to the corresponding to the combination of $O_1$ and $O_2$.

The most important thing to remark in this combination method is that it is a mixture of Classical Statistics based on likelihood functions and Bayesian Statistics. When we have a probabilistic 'a priori' information then Bayes Theorem is obtained. When we do not have 'a priori' information a likelihood function or its corresponding possibility is considered. When we have an 'a priori' information consisting on a convex set of probabilities, then Bayes Theorem is applied to each individual probability but, at the same time, it is defined a likelihood about the possible probabilities. Then we are using at the same time Bayes Theorem and likelihood functions, the first is applied to transform probabilities, the second to change our belief about what is the true probability. The following example illustrates these ideas.

*Example.-* Assume as above that we have two urns $U_1$ and $U_2$ with red and black balls:

| $U1$ | 99 red | 1 black |
|------|--------|---------|
| $U2$ | 1 red  | 99 black |

Consider that we pick up two balls with replacement from the same urn and the events are denoted as follows:

- B1 The first ball is black
- R1 The first ball is red
- B2 The second ball is black
- R2 The second ball is red

For the color of the two balls we have an 'a priori' information with two extreme probabilities, one for each urn:

|       | $R1 \cap R2$ | $R1 \cap B2$ | $B1 \cap R2$ | $B1 \cap B2$ |
|-------|--------------|--------------|--------------|--------------|
| $p_1$ | 0.9801       | 0.0099       | 0.0099       | 0.0001       |
| $p_2$ | 0.0001       | 0.0099       | 0.0099       | 0.9801       |

Assume now that we observe the colour of the first ball: red. This defines the following likelihood

|     | $R1 \cap R2$ | $R1 \cap B2$ | $B1 \cap R2$ | $B1 \cap B2$ |
|-----|--------------|--------------|--------------|--------------|
| $l$ | 1            | 1            | 0            | 0            |

The combination of 'a priori' information and the likelihood is

|       | $R1 \cap R2$ | $R1 \cap B2$ | $B1 \cap R2$ | $B1 \cap B2$ |
|-------|--------------|--------------|--------------|--------------|
| $h_1$ | 0.9801       | 0.099        | 0            | 0            |
| $h_2$ | 0.0001       | 0.099        | 0            | 0            |

If we normalize the probabilities calculating the corresponding possibilities we get

|              | $R2$ | $B2$ | $\pi$ |
|--------------|------|------|-------|
| $p_1(.|R1)$  | 0.99 | 0.01 | 0.99  |
| $p_2(.|R1)$  | 0.01 | 0.99 | 0.01  |

Observe as we have transformed each probability distribution according to Bayes rule. But also, the combination defines a possibility about which is the true probability (or what is equivalent: which is the true urn). Note also that here the probabilities for the second ball are the same as before the first ball is observed. However, knowing that the first ball is red defines a likelihood about which is the true urn, that changes our belief about the colour of the second ball. The integration of conditional probabilities and possibilities by using Choquet integral produces the following intervals:

$$R2 \quad [0.9801, 0.9900]$$
$$B2 \quad [0.0100, 0.0199]$$

These intervals incorporate not only the changes on conditonal probabilities but also our chages on belief about the urns, that is, the bayesian updating and the likelihood information.

**Acknowledgments**

We are indebted to L.M. de Campos, M. Delgado and M.T. Lamata for their help in completing this work. We are also very grateful to Ph. Smets by his valuable and useful comments.

This research has been supported by the Commission of European Communities under Project DRUMS (Esprit B.R.A. 3085).

**References**

Campos L.M. de, Lamata M.T., Moral S. (1988) Logical connectives for combining fuzzy measures. In: Methodologies for Intelligent Systems (Z.W. Ras, L. Saitta, eds.) Elsevier (New York) 11,18.

Campos L.M. de, Lamata M.T., Moral S.(1990) The concept of conditional fuzzy measure. International Journal of Intelligent Systems 5, 237-246.

Choquet G. (1953/54) Theorie of capacities. Ann. Inst. Fourier 5, 131-292.

Dempster A.P. (1967) Upper and lower probabilities induced by a multivalued mapping. Ann. Math. Statist. 38, 325-339.

Dubois D., Prade H. (1988). Possibility Theory. An Approach to Computerized Processing of Information. Plenum Press (New York).

Fagin R., Halpern J.Y. (1990) A new approach to updating beliefs. Research Report RJ 7222. IBM Almaden Research Center.

Moral S., Campos L.M. de (1990) Updating uncertain information. Proceedings 3rd. IPMU Conference, Paris 1990, 452-454.

Pearl J. (1989) Reasoning with belief functions: a critical assessment. Tech. Rep. R-136. University of California, Los Angeles.

Shafer G. (1976) A Mathematical Theory of Evidence. Princeton University Press (Princeton).

Smets Ph. (1982) Possibilistic Inference from Statistical Data. Proceedings of the Second World Conference on Mathematics at the Service of Man (A. Ballester, eds.) 611-613.

Smets Ph. (1978) Un Modele Mathematico-Statistique Simulant le Processus du Diagnostic Medical. Doctoral Disertation. Universite Libre de Bruxelles, Bruxelles.

Smets Ph. (1981) Medical Diagnosis: Fuzzy Sets and Degrees of Belief. Fuzzy Sets and Systems 5, 259-266.

Sugeno M. (1974) Theory of Fuzzy Integrals and its Applications. Ph. D. Thesis, Tokyo Institute of Technology.

Tessen B. (1989) Interval Representation of Uncertainty in Artificial Intelligence. Ph. D. Thesis, Departament of Informatics, University of Bergen, Norway.

Zadeh L.A. (1978) Fuzzy sets as a basis for a theory of possibility. Fuzzy Sets and Systems 1, 3-28.